\documentclass[a4paper, 10pt, conference]{ieeeconf}      
\usepackage{FG2025}
\FGfinalcopy 

\IEEEoverridecommandlockouts                              
\usepackage{cite}
\usepackage{amsmath,amssymb,amsfonts}
\usepackage{algorithmic}
\usepackage{graphicx}
\usepackage{graphics}
\usepackage{textcomp}
\usepackage{xcolor}
\usepackage{subfigure}
\usepackage{array}
\usepackage{multirow, makecell}
\usepackage{multicol}
\usepackage{booktabs} 
\usepackage{url}

\def\FGPaperID{137} 

\title{\LARGE \bf
MultiSensor-Home: A Wide-area Multi-modal Multi-view Dataset for Action Recognition and Transformer-based Sensor Fusion
}

\author{\parbox{18cm}{\centering
    {\large Trung Thanh Nguyen$^{1, 2}$, Yasutomo Kawanishi$^{2,1,3}$, Vijay John$^{2}$, Takahiro Komamizu$^{3, 1}$, and Ichiro Ide$^{1, 3}$}\\
    {\normalsize
    $^1$Graduate School of Informatics, Nagoya University, Nagoya, Aichi 464-8601, Japan\\
    $^2$Guardian Robot Project, Information R\&D and Strategy Headquarters, RIKEN, Seika, Kyoto 619-0288, Japan\\
    $^3$Center for Artificial Intelligence, Mathematical and Data Science, Nagoya University, Nagoya, Aichi 464-8601, Japan
    }}
}

\usepackage{fancyhdr}
\begin{document}

\ifFGfinal
\thispagestyle{empty}
\pagestyle{empty}
\else
\author{Anonymous FG2025 submission\\ Paper ID \FGPaperID \\}
\pagestyle{plain}
\fi
\maketitle

\thispagestyle{fancy} 

\begin{abstract}
Multi-modal multi-view action recognition is a rapidly growing field in computer vision, offering significant potential for applications in surveillance. 
However, current datasets often fail to address real-world challenges such as wide-area distributed settings, asynchronous data streams, and the lack of frame-level annotations. 
Furthermore, existing methods face difficulties in effectively modeling inter-view relationships and enhancing spatial feature learning.
In this paper, we introduce the MultiSensor-Home dataset, a novel benchmark designed for comprehensive action recognition in home environments, and also propose the Multi-modal Multi-view Transformer-based Sensor Fusion (MultiTSF) method. 
The proposed MultiSensor-Home dataset features untrimmed videos captured by distributed sensors, providing high-resolution RGB and audio data along with detailed multi-view frame-level action labels. 
The proposed MultiTSF method leverages a Transformer-based fusion mechanism to dynamically model inter-view relationships. 
Furthermore, the proposed method integrates a human detection module to enhance spatial feature learning, guiding the model to prioritize frames with human activity to enhance action the recognition accuracy.
Experiments on the proposed MultiSensor-Home and the existing MM-Office datasets demonstrate the superiority of MultiTSF over the state-of-the-art methods. 
Quantitative and qualitative results highlight the effectiveness of the proposed method in advancing real-world multi-modal multi-view action recognition.

\end{abstract}

\section{Introduction}
\label{sec:introduction}
Action recognition is a critical area of research in computer vision, with applications spanning surveillance~\cite{khan2024human}, robotics~\cite{voronin2021action}, and video content analysis~\cite{pareek2021survey}. 
Traditional single-view action recognition approaches~\cite{gupta2021quo, sun2022human, kong2022human} are constrained by their reliance on a single field-of-view, resulting in incomplete contextual understanding and misclassification, particularly in cases of occlusion or partial visibility of actions. 
These limitations have driven the adoption of multi-view systems~\cite{olagoke2020literature}, which allow actions to be observed from multiple field-of-views, enabling a more comprehensive and accurate understanding of human activities.

\begin{figure}[t]
    \centering
    \includegraphics[width=0.485\textwidth]{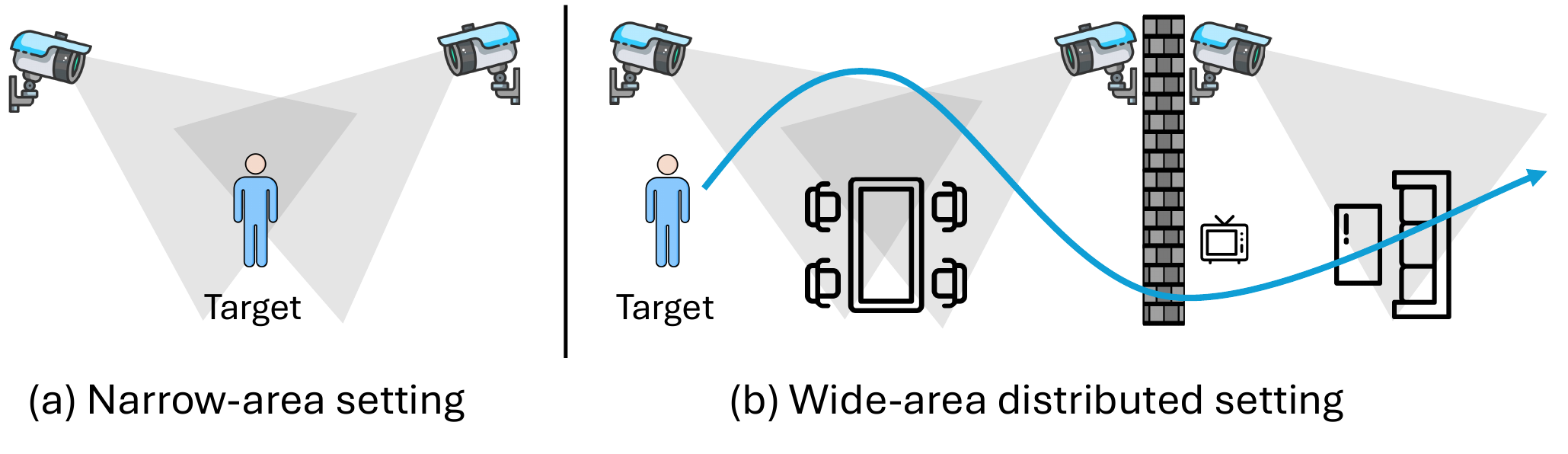}
    \caption{Configuration of multi-view settings. 
    (a) Multiple sensors capturing the same area. 
    (b) Multiple sensors capturing different areas, which is the environment targeted in this study.}
    \label{fig:multi_view_settings}
    \vspace{-5pt}
\end{figure}

Multi-view action recognition integrates visual data from multiple cameras to exploit complementary perspectives and capture the full spatial context of an action. 
While most existing studies~\cite{wang2021continuous, bai2020collaborative, shah2023multi, 10.1007/978-3-030-58583-9_26} focus on sensor setups in narrow-area coverage (Figure~\ref{fig:multi_view_settings}(a)), they often fail to generalize to complex, real-world scenarios where actions occur over wider areas and across viewpoints (Figure~\ref{fig:multi_view_settings}(b)). 
In a wide-area coverage distributed setting, the spatial dispersion of sensors introduces additional challenges, such as targets moving across multiple views and the need to maintain consistent tracking of actions across diverse perspectives.
These issues are compounded by the requirement for efficient fusion of complementary multi-view information while minimizing redundancy.

Recent advancements in multi-modal action recognition have highlighted the benefits of integrating diverse sensory inputs, such as audio and video~\cite{nagrani2021attention, gao2020listen}. 
Existing multi-modal approaches have been limited to single-view setting, and there remains a lack of methods addressing the challenges of integrating multi-modal data in multi-view configurations. 
On the other hand, existing multi-modal multi-view datasets designed for wide-area distributed setting are limited by several factors. 
Most notably, these datasets often provide only weak video sequence-level labels, which limits their usability for tasks requiring fine-grained spatial and temporal analysis. 
For instance, the MM-Office~\cite{yasuda2022multi} and MM-Store~\cite{yasuda2024guided} datasets lack detailed frame-level annotations for majority of the sequences, making them unsuitable for strongly supervised learning approaches. 

\begin{table*}[th]
\centering
\caption{Comparison of the proposed MultiSensor-Home dataset with existing multi-modal multi-view action recognition datasets.  \\
The proposed dataset offers high-resolution data with frame-level annotations.
} 
\resizebox{\textwidth}{!}
{
  \begin{tabular}{c|l|c|l|c|r|c|c|c|l}
    \toprule
    Setting & Dataset Name & Year  & \multirowcell{1}[0pt][c]{Modality} & \#Views & \multirowcell{1}[0pt][c]{\#Videos} & Avg. Duration & \#Classes   & \multirowcell{1}[0pt][c]{Resolution} & \multirowcell{1}[0pt][c]{Annotation} \\
    \midrule
    \multirowcell{4}[0pt][c]{Narrow\\Area} & NW-UCLA~\cite{wang2014cross} & 2014 &  RGB+D & 3 &  1,494 & 10 seconds & \hspace{2pt} 10 & ~~640 $\times$ ~~480 & Video-level \\
    & NTU RGB+D~\cite{shahroudy2016ntu} & 2016  & RGB+D & 3 & 56,880 & \hspace{2pt} 5 seconds & \hspace{2pt} 60 & 1,920 $\times$ 1,080 & Video-level \\
    & NTU RGB+D 120~\cite{liu2019ntu} & 2019 & RGB+D & 3 & 114,480 & \hspace{2pt} 5 seconds & 120 & 1,920 $\times$ 1,080 & Video-level \\
    & Toyota Smarthome~\cite{das2019toyota} & 2020 & RGB+D & 7 & 16,115 & 12 seconds & \hspace{2pt} 31 & ~~640 $\times$ ~~480 & Video-level \\ 
    \midrule
    \multirowcell{3}[0pt][c]{Wide\\Area} & MM-Office~\cite{yasuda2022multi} & 2022 &  RGB+Audio & 4 & 1,760 & 60 seconds & \hspace{2pt} 12 & 2,560 $\times$ 1,440 & Video-level \\ 
    & MM-Store~\cite{yasuda2024guided} & 2024 & RGB+Audio & 6 & 2,970 & 60 seconds & \hspace{2pt} 18 & 3,840 $\times$ 2,160 & Video-level \\
    & MultiSensor-Home (Proposed) & 2025 & RGB+Audio & 5 & 2,555 & 80 seconds & \hspace{2pt} 16 & 4,000 $\times$ 3,000 & Frame-level \\
    
    \bottomrule
  \end{tabular}
  \label{table:comparison_dataset}
}
\end{table*}

To address these challenges, we introduce a new benchmark, the MultiSensor-Home dataset, which provides untrimmed videos that include multiple actions captured across multiple views with detailed frame-level annotations. 
These videos are recorded using distributed sensors covering a wide-area, including audio and RGB image modalities. 
The proposed dataset also encompasses diverse scenarios, such as variations in time of day, clothing, and environmental conditions, making it a robust resource for studying real-world action recognition.
We also propose the Multi-modal Multi-view Transformer-based Sensor Fusion (MultiTSF) method, which is applicable to both narrow-area and wide-area distributed settings.
MultiTSF utilizes a Transformer-based attention mechanism to dynamically model inter-view relationships and capture temporal dependencies across multiple sensors. 
Additionally, to address wide-area distributed scenarios where targets move across dispersed views, we integrate a Human Detection Module to enhance spatial feature learning. This module enables the model to prioritize frames and views with human activity, which is crucial for reducing redundancy and concentrating on actionable data.

The key contributions of this study are as follows:
\begin{itemize}
    \item \textbf{MultiSensor-Home Dataset.} We introduce a multi-modal multi-view dataset with fine-grained multi-view frame-level action labels. 
    This dataset addresses the limitations of existing datasets by incorporating varied environmental conditions, diverse action scenarios, and synchronized audio-visual data. 
    \item \textbf{MultiTSF Method.} We propose a method for multi-modal multi-view action recognition that combines audio and visual inputs using a Transformer-based sensor fusion mechanism to model inter-view relationships dynamically.
    To support spatial feature learning in MultiTSF, we introduce a Human Detection Module that generates pseudo-ground-truth annotations for human presence. 
    This guides the model to prioritize actionable frames, which are frames containing human activity relevant to action recognition.
    \item \textbf{Extensive Evaluation.} 
    We evaluate MultiTSF on the proposed MultiSensor-Home dataset and the existing MM-Office dataset~\cite{yasuda2022multi}, showing significant improvements over state-of-the-art methods. 
    Quantitative and qualitative results demonstrate the effectiveness of the proposed method.
\end{itemize}

The remainder of this paper is structured as follows: Section~\ref{sec:related_work} reviews related work. Sections~\ref{sec:mmhome_dataset} and ~\ref{sec:methodology} introduce the proposed MultiSensor-Home dataset and the MultiTSF method, respectively. Section~\ref{sec:performance_evaluation} details the experimental results and analysis. Finally, Section~\ref{sec:conclusion} concludes the paper.

\section{Related Work}
\label{sec:related_work}
\begin{figure*}[th]
    \centering
    \begin{minipage}[t]{0.36\textwidth}
        \centering
        \includegraphics[width=\textwidth]{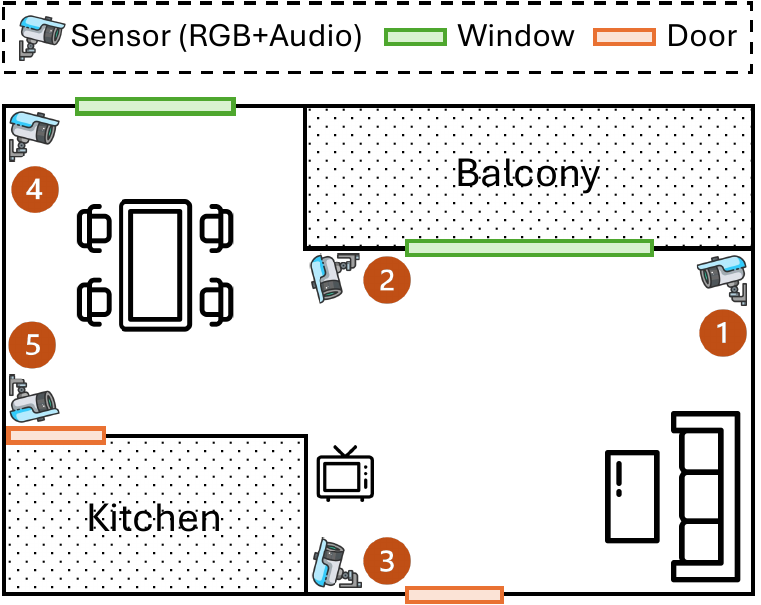}
        \caption{Room layout illustrating the placement of multi-view sensors in the proposed MultiSensor-Home dataset.}
        \label{fig:room_layout}
    \end{minipage}%
    \hfill
    \begin{minipage}[t]{0.60\textwidth}
        \centering
        \includegraphics[width=\textwidth]{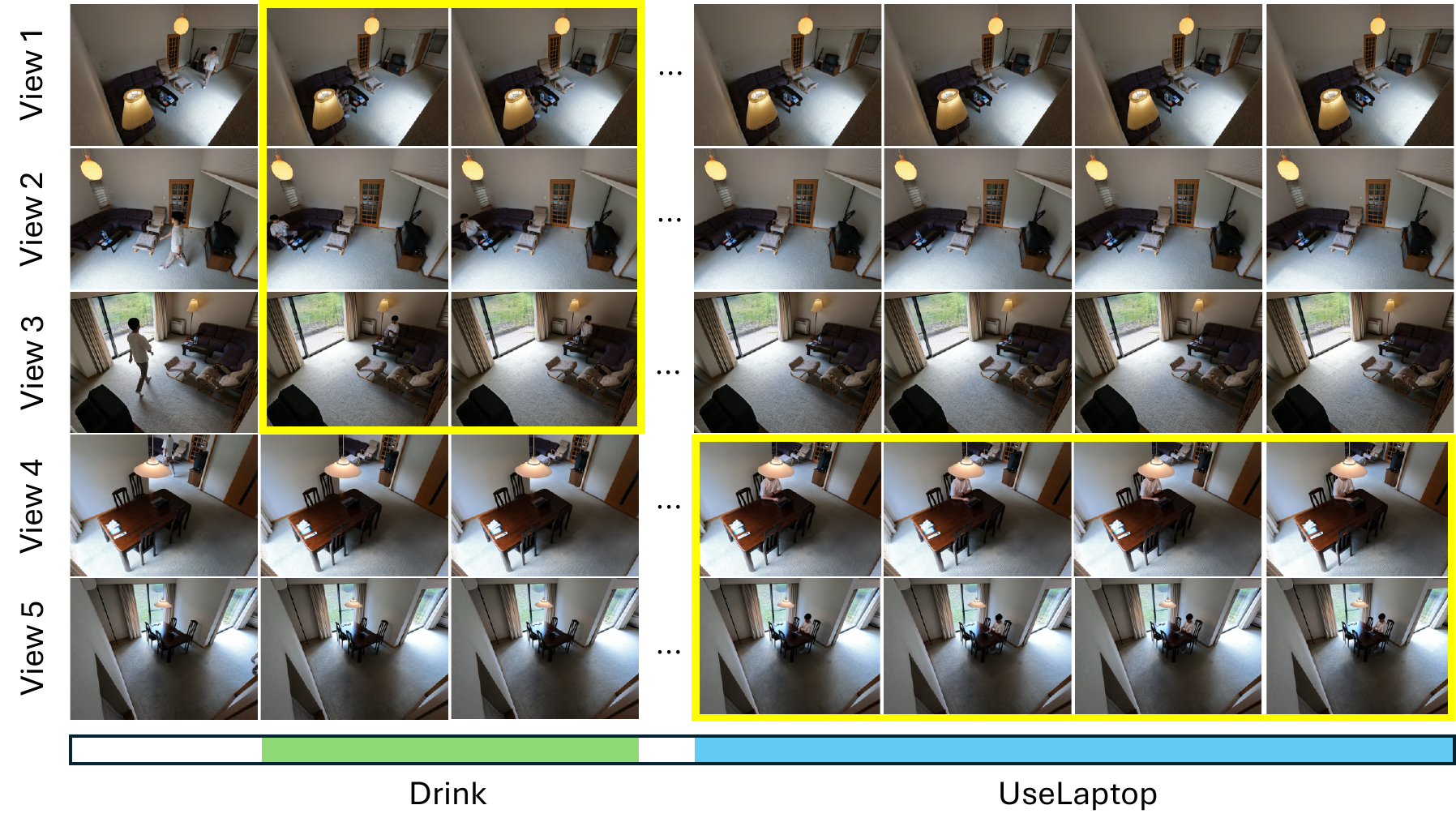}
        \caption{Example from the proposed MultiSensor-Home dataset showcasing actions captured from multiple views.}
        \label{fig:dataset_example}
    \end{minipage}
\end{figure*}


\subsection{Multi-modal Multi-view Datasets}
\label{subsec:dataset}
The advancement of action recognition techniques is enabled by the development of multi-modal multi-view datasets.
These datasets provide diverse perspectives and complementary modalities, allowing for a more comprehensive understanding of human actions.
One of the earliest publicly available datasets, the NorthWestern-UCLA Multi-view Action 3D (NW-UCLA) dataset~\cite{wang2014cross} utilizes narrow-area object-centered sensors to record RGB+D data (RGB color and Depth information) from three camera views. 
Following that, the Nanyang Technological University (NTU) RGB+D~\cite{shahroudy2016ntu} and its extension NTU RGB+D 120~\cite{liu2019ntu} datasets represent significant advancements by providing large-scale multi-modal multi-view recordings. 
They offer three synchronized views and cover up to 120 action classes, making them among the most comprehensive resources for action recognition. 
However, these datasets focus on controlled, narrow-area setting, with trimmed videos of short duration, which limit their applicability to real-world action recognition tasks.
To address the limitations of controlled settings, the Toyota Smarthome dataset~\cite{das2019toyota} introduced real-world home environments with RGB+D recordings captured from seven views in a dining room, kitchen, and living room. 
However, it lacks synchronized views, as each action is clipped per view, limiting its focus to a narrow-area scene.

Recently, Yasuda et al. introduced the MM-Office~\cite{yasuda2022multi} and MM-Store~\cite{yasuda2024guided} datasets, which feature RGB+Audio recordings captured in office and convenience store environments, respectively. 
These datasets utilize distributed sensors to capture a wide-area field-of-view, where camera views partially overlap each other. 
This setup reflects real-world scenarios, enabling the comprehensive study of human actions across a wide range of activities. 
While these dataset represent significant milestones in wide-area multi-view action recognition, they offer only video sequence-level labels or limited frame-level annotations confined to the test set.
Thus, their use is limited to weakly supervised learning tasks. 

To address these gaps, we propose the MultiSensor-Home dataset, a wide-area multi-modal multi-view dataset featuring untrimmed recordings captured across varying times of day (i.e., morning, afternoon, and evening). 
It also provides detailed multi-view frame-level labels, enabling its use in tasks such as strongly supervised action recognition at the frame
level and temporal action recognition. 
Table~\ref{table:comparison_dataset} provides a comparison of the proposed MultiSensor-Home dataset with existing multi-modal multi-view action recognition datasets.

\subsection{Multi-modal Multi-view Action Recognition}
\label{subsec:recent_research}
Recent advancements in action recognition have focused on leveraging complementary information from multiple data modalities and camera viewpoints to enhance robustness and accuracy in human action recognition tasks. 
Most existing methods are constrained to setups with cameras, primarily due to the availability of narrow-area datasets. 
Skeleton-based approaches utilize detailed annotations to model human body dynamics across multiple views~\cite{shi2020decoupled, 9710612, zhang2019view}. 
However, these methods often rely on specific features that struggle to generalize in dynamic or unconstrained environments.
Conversely, recent studies have explored image-based multi-view action recognition.
Techniques such as supervised contrastive learning~\cite{shah2023multi} have been used to enhance feature robustness to viewpoint variations, while unsupervised representation learning~\cite{10.1007/978-3-030-58583-9_26} has been employed to create embeddings robust to changes in perspective.

To enable wide-area action recognition in real-world environments, recent works have introduced distributed sensor systems and advanced fusion strategies. 
Yasuda et al. proposed MultiTrans~\cite{yasuda2022multi}, a method designed to integrate data from distributed sensors by modeling inter-sensor relationships. 
However, it does not incorporate temporal dynamics, which are crucial for capturing the sequential nature of actions in action recognition tasks.
Similarly, an extended version of MultiTrans with Guided-MELD~\cite{yasuda2024guided} addresses the challenges of fragmented sensor observations by distilling redundant information and supplementing missing sensor data to create comprehensive event representations.
On the other hand, John et al.~\cite{john2024frame} introduced a weakly supervised latent embedding model that uses view-specific latent embeddings for downstream frame-level action recognition and detection tasks. 
Nguyen et al. proposed MultiASL~\cite{nguyen2024action}, which addresses the lack of frame-level labels by introducing an Action Selection Learning (ASL) mechanism. 
This mechanism leverages video sequence-level annotations to generate pseudo-frame-level labels for training the network.

Although these methods have achieved significant accuracy in multi-modal multi-view action recognition, their sensor fusion strategies often do not adequately address view invariance or incorporate temporal dynamics effectively.
To improve the accuracy, we propose MultiTSF method, which introduces a Transformer-based sensor fusion mechanism. 
Unlike MultiTrans, which focuses on sensor relationships at the video sequence-level, 
MultiTSF models the importance of sensor-level data at the frame-level by incorporating spatiotemporal features, ensuring a more detailed and context-aware representation of action dynamics.
Additionally, we introduce a Human Detection Module to enhance spatial feature learning to generate pseudo-ground-truth annotations for frames containing human presence, thereby improving the ability of the model to effectively capture relevant spatial and temporal features.


\section{MultiSensor-Home Dataset}
\label{sec:mmhome_dataset}
\begin{table}[t]
\centering
\caption{Action Classes in the proposed MultiSensor-Home Dataset. 
``\#Events'' indicates the number of occurrences.}
\label{table:action_classes}
{
\begin{tabular}{l|l|c}
\toprule
{Class}              & {Description}                  & {\#Events} \\
\midrule
AdjustAC                          & Adjusting the air conditioning unit   & 39                         \\
Clean                             & General cleaning activity             & 26                         \\
CleanVacuum                       & Cleaning using a vacuum cleaner       & 48                         \\
OpenCurtain                       & Opening the curtain                   & 38                         \\
CloseCurtain                      & Closing the curtain                   & 39                         \\
Drink                             & Drinking from a cup or bottle         & 51                         \\
Eat                               & Eating food                           & 48                         \\
Enter                             & Entering the room                     & 70                         \\
Exit                              & Exiting the room                      & 88                         \\
ReadBook                          & Reading a book                        & 64                         \\
Sitdown                           & Sitting down                          & \hspace{-3pt}247                         \\
Standup                           & Standing up                           & \hspace{-3pt}161                         \\
TurnOnLamp                        & Turning on the lamp                   & 57                         \\
TurnOffLamp                       & Turning off the lamp                  & 52                         \\
UseLaptop                         & Using a laptop computer               & \hspace{-3pt}196                         \\
UsePhone                          & Using a phone                         & \hspace{-3pt}110                         \\
\bottomrule
\end{tabular}
\label{table:action_classes}
}
\end{table}

\begin{figure*}[t]
    \centering
    \includegraphics[width=0.95\textwidth]{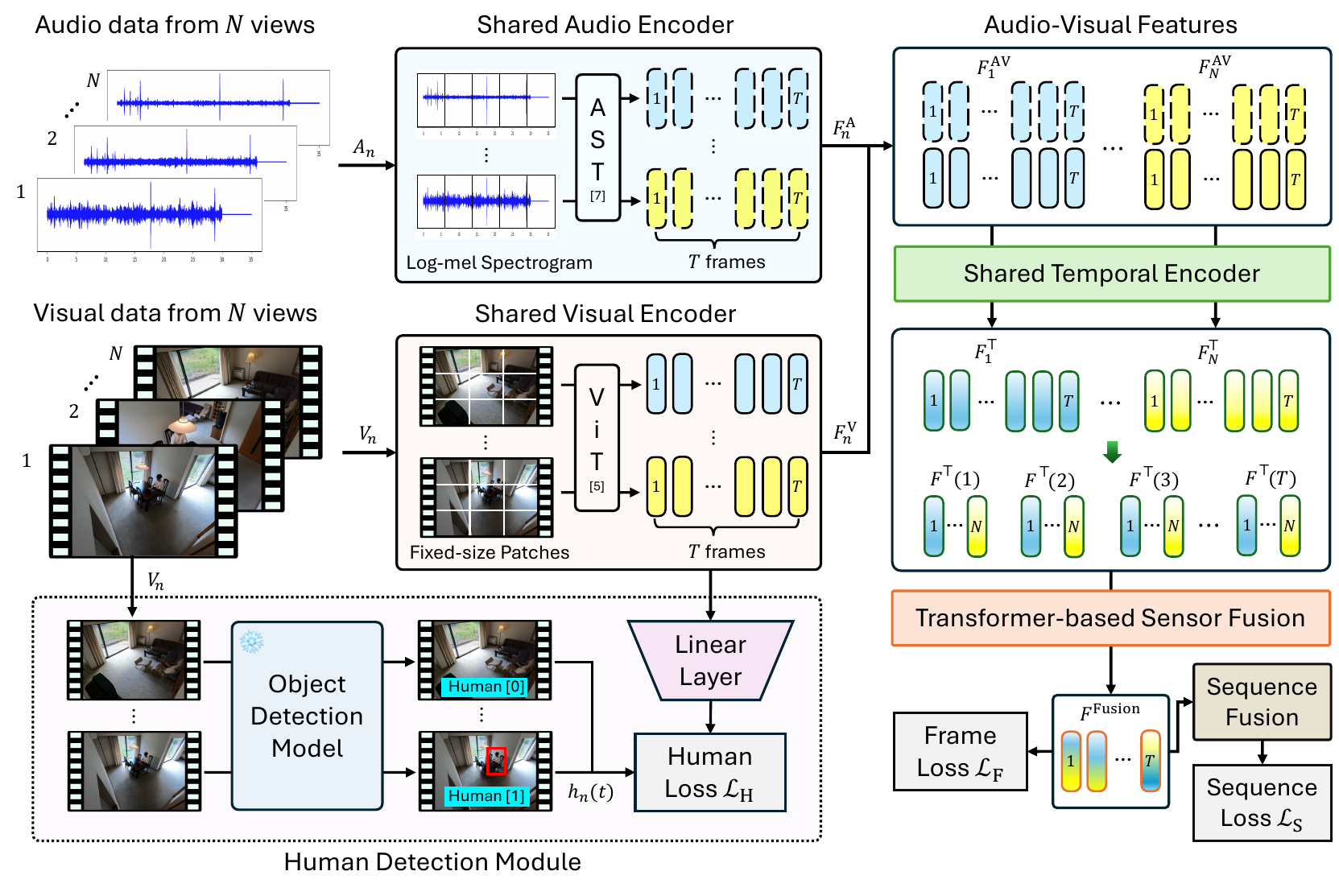}
    \caption{Overview of the proposed MultiTSF method. It consists of: (1) Multi-modal Feature Extraction using Shared Audio Encoder and Shared Visual Encoder to extract discriminative features, (2) Human Detection Module to detect human presence and generate pseudo-ground-truth labels, and (3) Temporal Modeling and Transformer-based Fusion to capture temporal dependencies and integrate spatiotemporal features for action recognition.}
    \label{fig:methodology}
\end{figure*}

We introduce the MultiSensor-Home dataset, a comprehensive benchmark for realistic multi-modal multi-view action recognition in indoor home environments. 
The dataset features untrimmed videos captured using five synchronized cameras strategically placed across a wide-area distributed setting, as illustrated in Figure~\ref{fig:room_layout}, where a single person performs actions within the room.
This multi-view configuration enables recording actions from diverse perspectives, ensuring comprehensive coverage of spatial dynamics within the environment. 
Each video includes RGB and audio modalities, recorded at a 4K resolution of $4,000 \times 3,000$ pixels and a frame rate of 30 frames per second, offering high-quality multi-modal data for action recognition tasks.
Moreover, MultiSensor-Home captures actions performed under diverse conditions, including different times of day, varying clothing styles, and natural variations in activity settings.
The dataset provides detailed multi-view frame-level annotations, enabling fine-grained spatial and temporal analysis. 
Figure~\ref{fig:dataset_example} showcases examples of actions captured from multiple views and Table~\ref{table:action_classes} presenting the action classes.

\section{Multi-modal Multi-view \\ Transformer-based Sensor Fusion}
\label{sec:methodology}
In this section, we propose the MultiTSF method for multi-modal multi-view action recognition. 
The problem definition and an overview of the proposed method are provided in Section~\ref{sub_sec:proposed_method}, followed by detailed explanations of each key component in Sections~\ref{sub_sec:multi_modal_feature_extraction}, ~\ref{sub_sec:human_detection_module}, and~\ref{sub_sec:temporal_modeling_Transformer_based_fusion}.

\subsection{Proposed Method}
\label{sub_sec:proposed_method}
\subsubsection{Problem Definition}
MultiTSF aims to predict action sequences (i.e., multi-label in the video) from the input multi-modal data captured by multi-view sensors (views). 
The input consists of a set of audio data \(A = \{A_1, A_2, \dots, A_N\}\) and corresponding video data \(V = \{V_1, V_2, \dots, V_N\}\), captured by \(N\) distributed sensors. 
Each audio data \(A_n \in \mathbb{R}^{T \times F}\) is represented as a spectrogram, where \(n \in \{1, 2, \dots, N\}\), \(T\) is the number of frames, and \(F\) is the number of frequency bins. Similarly, video data are represented as \(V_n \in \mathbb{R}^{T \times D \times H \times W}\), where \(T\) is the number of video frames, \(D\) is the number of channels, and \(H\) [pixels] and \(W\) [pixels] represent the height and width of each frame, respectively. 
The objective is to extract meaningful audio-visual and temporal features to predict multi-label action outputs. 
Frame-level predictions are defined as \(L_t \in \{0, 1\}^C\) for each frame \(t \in \{1, 2, \dots, T\}\), while sequence-level predictions are represented as \(L \in \{0, 1\}^C\), where \(C\) is the total number of action classes.

\subsubsection{Overview of MultiTSF}
To achieve robust performance in multi-modal multi-view action recognition, the proposed method integrates several key components, as illustrated in Figure~\ref{fig:methodology}.
The \textit{Multi-modal Feature Extraction} module extracts discriminative features from audio and visual streams. 
The \textit{Human Detection Module} employs an object detection model to detect frames with human presence and generate pseudo-ground-truth labels, guiding the model to effectively learn human activity within the visual features. 
To capture temporal dependencies and integrate multi-view spatiotemporal features, the \textit{Temporal Modeling and Transformer-based Fusion} component leverages a Transformer-based attention mechanism. Finally, the \textit{Learning Objectives} module optimizes the framework using frame-level, sequence-level, and human loss functions, enhancing spatial and temporal understanding for accurate action recognition.

\subsection{Multi-modal Feature Extraction}
\label{sub_sec:multi_modal_feature_extraction}
In this module, we employ a Shared Audio Encoder and a Shared Visual Encoder to process audio and visual input data. 
These encoders use the same model parameters for all input views, ensuring consistent and efficient feature extraction.
\subsubsection{Shared Audio Encoder}
Each raw audio signal \(A_n\) from the \(n\)-th view (\(n \in \{1, 2, \dots, N\}\)) is transformed into a log-mel spectrogram, which captures the temporal and frequency characteristics of the audio. 
These spectrograms are then processed using a shared Audio Spectrogram Transformer (AST) model~\cite{gong2021ast} to extract discriminative features.
AST divides the spectrogram into overlapping patches, projects them into embeddings, and processes the sequence of embeddings through Transformer layers. The resulting audio features are represented as:
\begin{equation}    
F_n^{\mathsf{A}} = f_a(A_n), \quad F_n^{\mathsf{A}} \in \mathbb{R}^{T \times D_{\mathsf{A}}},
\end{equation}
where \(f_a(\cdot)\) represents the AST model operation, and \(D_{\mathsf{A}}\) is the dimensionality of the extracted audio features.

\subsubsection{Shared Visual Encoder}
To extract spatial features from each video frame in each video \(V_n\), we use a shared 
Vision Transformer (ViT) model~\cite{dosovitskiy2020image}. 
ViT splits each video frame into non-overlapping patches of size \(P \times P\) [pixels], flattens these patches, and projects them into embeddings. 
These embeddings are then processed by Transformer layers to capture spatial dependencies. 
The visual features for the \(n\)-th view are represented~as:
\begin{equation}
    F_n^{\mathsf{V}} = f_v(V_n), \quad F_n^{\mathsf{V}} \in \mathbb{R}^{T \times D_{\mathsf{V}}},
\end{equation}
where \(f_v(\cdot)\) represents the ViT model operation, and \(D_{\mathsf{V}}\) is the dimensionality of the extracted visual features.

\subsubsection{Audio-Visual Features}
The extracted audio features \(F_n^{\mathsf{A}}\) and visual features \(F_n^{\mathsf{V}}\) are concatenated at the frame level to form combined audio-visual features. 
For each frame \(t \in \{1, 2, \dots, T\}\), the audio-visual feature is computed as:
\begin{equation}
    F_n^{\mathsf{AV}}(t) = [F_n^{\mathsf{A}}(t); F_n^{\mathsf{V}}(t)],
\end{equation}
where \([ \cdot ; \cdot ]\) denotes concatenation along the feature dimension. The combined sequence of audio-visual features for the \(n\)-th view is represented as $F_n^{\mathsf{AV}} \in \mathbb{R}^{T \times D_{\mathsf{AV}}},$ where \(D_{\mathsf{AV}} = D_{\mathsf{A}} + D_{\mathsf{V}}\). These fused features are subsequently passed to the temporal modeling and fusion stages for further processing.

\subsection{Human Detection Module}
\label{sub_sec:human_detection_module}

To enhance spatial feature learning, we introduce a Human Detection Module that detects human presence in video frames and generates pseudo-ground-truth labels for the Human Loss function.
This module operates independently of the main framework and processes visual data from multiple views.
For each video \(V_n\) from the \(n\)-th view (\(n \in \{1, 2, \dots, N\}\)), the frames are passed through the Object Detection Model (i.e., You Only Look Once (YOLO) v10 model~\cite{wang2024yolov10}), which outputs a binary indicator for each frame \(t \in \{1, 2, \dots, T\}\) as:
\begin{equation}
h_n(t) = 
\begin{cases} 
1, & \text{if a human is detected in frame } t, \\
0, & \text{otherwise}.
\end{cases}    
\end{equation}
The binary outputs \(h_n(t)\) for all frames serve as pseudo-ground-truth labels to supervise the learning of human-related spatial features. 
These labels guide the model to focus on frames with human activity and ignore irrelevant frames.
During training, they are used in the Human Loss function, which is detailed in the \textit{Learning Objectives} section.

\subsection{Temporal Modeling and Transformer-based Fusion}
\label{sub_sec:temporal_modeling_Transformer_based_fusion}

\subsubsection{Shared Temporal Encoder}
We employ a Shared Temporal Encoder to capture temporal dependencies in the extracted audio-visual features. 
The input to the encoder is the concatenated audio-visual features $F_n^{\mathsf{AV}}$ from each view \( n \in \{1, 2, \dots, N\} \). 
The Shared Temporal Encoder uses a self-attention mechanism~\cite{vaswani2017attention} to model the temporal relationships across frames within a view. Specifically, the self-attention mechanism computes the weighted interactions between frames as:
\begin{equation}
\text{Attention}(Q, K, V) = \text{softmax}\left( \frac{QK^\top}{\sqrt{d_k}} \right) V,
\end{equation}
where \( Q, K, V \in \mathbb{R}^{T \times d_k} \) are the query, key, and value matrices derived from \( F_n^{\mathsf{AV}} \), and \( d_k \) is the dimensionality of the query and key spaces. 
The output of this encoder for each view \( n \) is a set of temporally enhanced features as:
\begin{equation}
F_n^{\mathsf{T}} = \left[ F_n^{\mathsf{T}}(1), F_n^{\mathsf{T}}(2), \dots, F_n^{\mathsf{T}}(T) \right] \in \mathbb{R}^{T \times D_{\mathsf{T}}},
\end{equation}
where \( D_{\mathsf{T}} \) is the dimensionality of the temporal features.

\subsubsection{Transfomer-based Sensor Fusion}
After extracting temporal features \( F_n^{\mathsf{T}} \) from all \( N \) views, the Transformer-based Sensor Fusion mechanism combines information from multiple views to generate a unified representation. 
For each frame \( t \in \{1, 2, \dots, T\} \), the temporal features from all \( N \) views are processed to capture inter-view relationships. 
Specifically, for a given frame \( t \), the temporal features from the \( N \) views are represented as:
\begin{equation}
    F^{\mathsf{T}}(t) = \left[ F_1^{\mathsf{T}}(t), F_2^{\mathsf{T}}(t), \dots, F_N^{\mathsf{T}}(t) \right] \in \mathbb{R}^{N \times D_{\mathsf{T}}}.
\end{equation}
These features are input to the self-attention mechanism to model the importance and relationships between views for the same frame \( t \). 
This mechanism allows the model to assign dynamic importance to different views based on their contributions to the frame's action recognition. 
The output is a fused feature for the frame $
F^{\mathsf{Fusion}}(t) \in \mathbb{R}^{D_{\mathsf{Fusion}}},$
where \( D_{\mathsf{Fusion}} \) is the dimensionality of the fused features. 
Repeating this process for all \( T \) frames yields the final sequence of fused features as:
\begin{equation}
F^{\mathsf{Fusion}} = \left[ F^{\mathsf{Fusion}}(1), F^{\mathsf{Fusion}}(2), \dots, F^{\mathsf{Fusion}}(T) \right],
\end{equation}
where $F^{\mathsf{Fusion}}  \in \mathbb{R}^{T \times D_{\mathsf{Fusion}}}$.

\subsection{Learning Objectives}
\label{sub:learning_objectvies}
We design the learning objectives to optimize spatial, temporal, and inter-view features for effective action recognition. 
Human Loss \( \mathcal{L}_{\text{H}} \) ensures focus on frames with human activity. 
Frame Loss \( \mathcal{L}_{\text{F}} \) and Sequence Loss \( \mathcal{L}_{\text{S}} \) address class imbalance at the frame-level and sequence-level, respectively.
The total loss function is computed as:
\begin{equation}
    \mathcal{L} = \beta_1 \mathcal{L}_{\text{H}} + \beta_2 \mathcal{L}_{\text{F}} + \beta_3 \mathcal{L}_{\text{S}},
\end{equation}
where \( \beta_1 \), \( \beta_2 \), and \( \beta_3 \) are hyperparameters that control the relative importance of each loss term. 

\subsubsection{Human Loss \( \mathcal{L}_{\mathrm{H}} \)} 
This loss ensures that the model effectively learns to identify whether a frame contains human activity across \( N \) views. 
For each view \( n \in \{1, \dots, N\} \), the Shared Visual Encoder extracts frame-level features. 
These features are processed through a linear layer to predict the probability \( \hat{h}_n(t) \) of human presence in frame \(t \in \{1, 2, \dots, T\}\). 
The Binary Cross Entropy loss is used to measure the discrepancy between the predicted probabilities \( \hat{h}_n(t) \) and the pseudo-ground-truth labels \( h_n(t) \) as:
\begin{equation}
\small
    \begin{aligned}
        \mathcal{L}_{\text{H}} = - \frac{1}{N T} \sum_{n=1}^{N} \sum_{t=1}^{T} 
        \Big[ & h_n(t) \log \hat{h}_n(t) \\
        & + \big( 1 - h_n(t) \big) \log \big( 1 - \hat{h}_n(t) \big) \Big].
    \end{aligned}
\end{equation}
%

\subsubsection{Frame Loss \( \mathcal{L}_{\mathrm{F}} \)}
This loss optimizes the model's ability to predict frame-level action classes. To address class imbalance, we employ a two-way loss function~\cite{kobayashi2023two}, which combines \textit{sample-wise} and \textit{class-wise} components as:
\begin{equation}
    \mathcal{L}_{\text{F}} = \mathcal{L}_{\text{F}}^{\mathsf{S}} + \alpha_1 \mathcal{L}_{\text{F}}^{\mathsf{C}},
\end{equation}
where \( \alpha_1 \) is a balancing parameter.

Here, the sample-wise loss $\mathcal{L}_{\text{F}}^{\mathsf{S}}$ for frames discriminates between positive and negative classes as:
\begin{equation}
\small
    \mathcal{L}_{\text{F}}^{\mathsf{S}} = \frac{1}{T} \sum_{t=1}^{T} \text{softplus} \left( 
    \log \sum_{n \in \mathcal{N}_t} e^{x_{\mathsf{S}_n}} 
    + \gamma_s \log \sum_{p \in \mathcal{P}_t} e^{-\frac{x_{\mathsf{S}_p}}{\gamma_s}} 
    \right),
\end{equation}
where $\mathcal{N}_t$ and $\mathcal{P}_t$ denote the index sets of negative and positive samples for frame $t$, respectively; 
$x_{\mathsf{S}_n}$ and $x_{\mathsf{S}_p}$ are the logits corresponding to negative and positive samples; and $\gamma_s$ is a temperature parameter.
The function $\text{softplus}(\cdot) = \log(1+\exp(\cdot))$ serves as a smooth approximation to the Rectified Linear Unit (ReLU) function.



Meanwhile, the class-wise loss $\mathcal{L}_{\text{F}}^{\mathsf{C}}$ addresses intra-class variations across frames as:
\begin{equation}
\small
    \mathcal{L}_{\text{F}}^{\mathsf{C}} = \frac{1}{C} \sum_{c=1}^{C} \text{softplus} \left( 
    \log \sum_{n \in \mathcal{N}_c} e^{x_{\mathsf{C}_n}} 
    + \gamma_c \log \sum_{p \in \mathcal{P}_c} e^{-\frac{x_{\mathsf{C}_p}}{\gamma_c}} 
    \right),
\end{equation}
where $\mathcal{N}_c$ and $\mathcal{P}_c$ represent the index sets of negative and positive samples within class $c$, respectively;
$x_{\mathsf{C}_n}$ and $x_{\mathsf{C}_p}$ are their corresponding logits, and $\gamma_c$ is a temperature parameter.

\subsubsection{Sequence Loss \( \mathcal{L}_{\mathrm{S}} \)}
This loss optimizes sequence-level predictions by aggregating temporal features across \( T \) frames. 
The fused sequence features are passed through a classification head, and the loss is calculated using the two-way loss~\cite{kobayashi2023two} as:
\begin{equation}
    \mathcal{L}_{\text{S}} = \mathcal{L}_{\text{S}}^{\mathsf{S}} + \alpha_2 \mathcal{L}_{\text{S}}^{\mathsf{C}},
\end{equation}
where Sample-wise loss $\mathcal{L}_{\text{S}}^{\mathsf{S}}$ and Class-wise loss $\mathcal{L}_{\text{S}}^{\mathsf{C}}$ are computed similarly to Frame Loss \( \mathcal{L}_{\text{F}} \) but applied to sequence-level predictions, and \( \alpha_2 \) is a balancing parameter. 


\section{Performance Evaluation}
\label{sec:performance_evaluation}

\subsection{Experimental Conditions}
\subsubsection{Data Preparation}
We evaluate the proposed method using the proposed MultiSensor-Home and the existing MM-Office~\cite{yasuda2022multi} datasets. 
We apply the iterative stratification strategy~\cite{sechidis2011stratification} to divide the data into training and test subsets, ensuring a balanced representation of all classes in these subsets. The MultiSensor-Home dataset is split in a 70:30 ratio, while the MM-Office dataset follows the splitting strategy outlined in~\cite{nguyen2024action}.

For the experiments, we extract a fixed number of \(T\) synchronized frames, where visual frames are sampled uniformly from the video at a fixed frame rate (i.e., 2.5 FPS) to ensure consistent temporal spacing, and corresponding audio segments are aligned based on the exact timestamps of these frames.
During the training phase, we generate a sequence of frame indices using uniform sampling with random perturbations, ensuring that the sequence spans the entire video while maintaining a fixed length of \(T\). This approach serves as a data augmentation technique, enhancing the robustness of the model by introducing variability during training.
For testing, we apply uniform sampling without perturbation to ensure consistency across all test runs.

\subsubsection{Evaluation Metrics}
Following~\cite{kobayashi2023two, nguyen2024action}, we evaluate the performance using the following two metrics:
\begin{itemize}
    \item \(\text{mAP}_{C}\) (macro-averaged metric): The mean average precision is calculated for each class and then averaged across all \textit{classes}. This serves as the primary metric for multi-label classification.    
    \item \(\text{mAP}_{S}\) (micro-averaged metric): The mean average precision is computed across all \textit{samples}. This is commonly used as a standard metric for single-label classification.
\end{itemize}

\subsubsection{Comparison Methods}
We compare the proposed MultiTSF method with several state-of-the-art approaches for multi-modal multi-view and video-based action recognition. 
Specifically, for multi-modal multi-view methods, we include MultiTrans~\cite{yasuda2022multi} and MultiASL~\cite{nguyen2024action}. 
For video-based action recognition, we evaluate against TimeSformer~\cite{bertasius2021space} and Video Vision Transformer (ViViT)~\cite{arnab2021vivit}, which are Transformer-based architectures that capture spatiotemporal relationships, and X-CLIP~\cite{ma2022x}, which extends Contrastive Language-Image Pretraining (CLIP)~\cite{radford2021learning} to video data.

\begin{table}[t]
\centering
\caption{Comparison of the proposed MultiTSF method with other methods on the MultiSensor-Home and MM-Office datasets. 
The best and second-best results are highlighted in \textbf{bold} and \underline{underlined} text, respectively.} 
\subtable[Results on the MultiSensor-Home dataset in sequence-level and frame-level settings.]
{
  \begin{tabular}{l|cc|cc}
    \toprule
    \multirowcell{2}[-2pt][l]{Method} & \multicolumn{2}{c}{Sequence-level} & \multicolumn{2}{c}{Frame-level}  \\
    \cmidrule(lr){2-5}
    & mAP${_C}$ & mAP${_S}$ & mAP${_C}$ & mAP${_S}$ \\
    \midrule
    
    \multicolumn{5}{c}{Uni-modal (Visual)} \\
    \midrule
    TimeSformer~\cite{bertasius2021space} & 50.37 & 71.02 & -- & -- \\
    ViViT~\cite{arnab2021vivit} & 43.14 & 67.37 & -- & -- \\
    X-CLIP~\cite{ma2022x} & 42.57 & 68.83 & -- & -- \\
    MultiTrans~\cite{yasuda2022multi} & \underline{57.59} & 76.09 & 60.77 & 75.78 \\
    MultiASL~\cite{nguyen2024action} & 55.91 & \underline{77.25} & \underline{63.24} & \underline{80.64} \\
    MultiTSF (Proposed) & \textbf{61.17} & \textbf{84.22} & \textbf{75.07} & \textbf{87.31} \\
    \midrule
    
    \multicolumn{5}{c}{Multi-modal (Audio + Visual)} \\
    \midrule
    MultiTrans~\cite{yasuda2022multi} & \underline{59.65} & \underline{77.60} & 61.40 & 78.07 \\
    MultiASL~\cite{nguyen2024action} & 58.58 & 77.43 & \underline{73.81} & \underline{85.38} \\
    MultiTSF (Proposed) & \textbf{64.48} & \textbf{87.91} & \textbf{76.12} & \textbf{91.45} \\
    \bottomrule
  \end{tabular}
  \label{table:main_experiments}
}
\hfill
\subtable[Results on the MM-Office dataset~\cite{yasuda2022multi} in sequence-level setting.]
{
  \begin{tabular}{l|cc|cc}
    \toprule
     \multirowcell{2}[-2pt][l]{Method} & \multicolumn{2}{c}{Uni-modal (Visual)} & \multicolumn{2}{c}{Multi-modal}  \\
    \cmidrule(lr){2-5}
     & mAP${_C}$ & mAP${_S}$ & mAP${_C}$ & mAP${_S}$ \\
    \midrule
    TimeSformer~\cite{bertasius2021space} & 69.68 & 79.26 & -- & -- \\
    ViViT~\cite{arnab2021vivit} & 73.25 & 83.05 & -- & -- \\
    X-CLIP~\cite{ma2022x} & 65.38 & 78.54 & -- & -- \\
    MultiTrans~\cite{yasuda2022multi} & 73.85 & 85.24 & 75.35 & 85.35 \\
    MultiASL~\cite{nguyen2024action} & \underline{81.13} & \underline{89.52} & \textbf{86.23} & \underline{92.97} \\
    MultiTSF (Proposed) & \textbf{81.71} & \textbf{91.23} & \underline{85.65} & \textbf{93.03} \\
    \bottomrule
  \end{tabular}
  \label{table:mm_office_results}
}
\vspace{-8pt}
\end{table}

\subsubsection{Models \& Hyperparameters}
We implement the proposed MultiTSF as detailed in Section~\ref{sec:methodology}\footnote{The source code is available at \url{https://github.com/thanhhff/MultiTSF}.}. 
For MultiSensor-Home and MM-Office datasets, we use \(T = 70\) and \(T = 50\), respectively, sampled at 2.5 FPS, based on the average video lengths in each dataset. 
The Shared Audio Encoder extracts 128-dimensional log-mel filterbank features using AST~\cite{gong2021ast}, while the Shared Video Encoder processes frames with a resolution of \(224 \times 224\) pixels using ViT~\cite{dosovitskiy2020image}. 
The Shared Temporal Encoder uses a two-layer Transformer encoder with four attention heads of 128 dimensions. 
Transformer-based Sensor Fusion uses a single layer with four attention heads of 128 dimensions. 
The hyperparameters controlling the relative importance of each loss term are set to 1 for simplicity.
Optimization is performed using Adaptive moment estimation (Adam)~\cite{kingma2014adam} with an initial learning rate of \(10^{-4}\), a weight decay of \(5.0 \times 10^{-4}\), and a batch size of 12 for 300 epochs. 
The learning rate is scheduled using Cosine Annealing~\cite{loshchilov2016sgdr} to adaptively decay over training.
All experiments are conducted on a machine with four Tesla V100-PCIE-32GB GPUs.

\subsection{Quantitative Results}
Tables~\ref{table:main_experiments} and~\ref{table:mm_office_results} present the results of the proposed MultiTSF compared with other methods on the MultiSensor-Home and MM-Office~\cite{yasuda2022multi} datasets, respectively.

On the MultiSensor-Home dataset (Table~\ref{table:main_experiments}), MultiTSF achieved the best performance across all metrics and settings. 
Specifically, MultiTSF significantly outperformed MultiTrans~\cite{yasuda2022multi} and MultiASL~\cite{nguyen2024action} in the multi-modal setting, achieving {64.48\%} of mAP${_C}$ and {87.91\%} of mAP${_S}$ in the sequence-level setting (i.e., where only video sequence-level labels were available) and {76.12\%} of mAP${_C}$ and {91.45\%} of mAP${_S}$ in the frame-level setting (i.e., where frame-level labels were available). 
In the uni-modal setting, MultiTSF also surpassed TimeSformer~\cite{bertasius2021space}, ViViT~\cite{arnab2021vivit}, and X-CLIP~\cite{ma2022x}, demonstrating its robustness even when audio features were excluded.

On the MM-Office dataset (Table~\ref{table:mm_office_results}), MultiTSF maintained its superior performance in the sequence-level setting. 
In the uni-modal visual configuration, MultiTSF achieved the highest mAP${_C}$ of {81.71\%} and mAP${_S}$ of {91.23\%}, outperforming all baselines, including MultiASL and MultiTrans. 
In the multi-modal setting, MultiTSF achieved competitive results with {93.03\%} of mAP${_S}$, surpassing MultiASL and demonstrating its ability to effectively integrate audio and visual modalities.

\subsection{Qualitative Results}
\begin{figure}
    \centering
    \subfigure[Multi-view inputs (top row) and attention heatmaps (bottom row) highlighting action-relevant regions.]
    {
    \centering
    \includegraphics[width=0.47\textwidth]{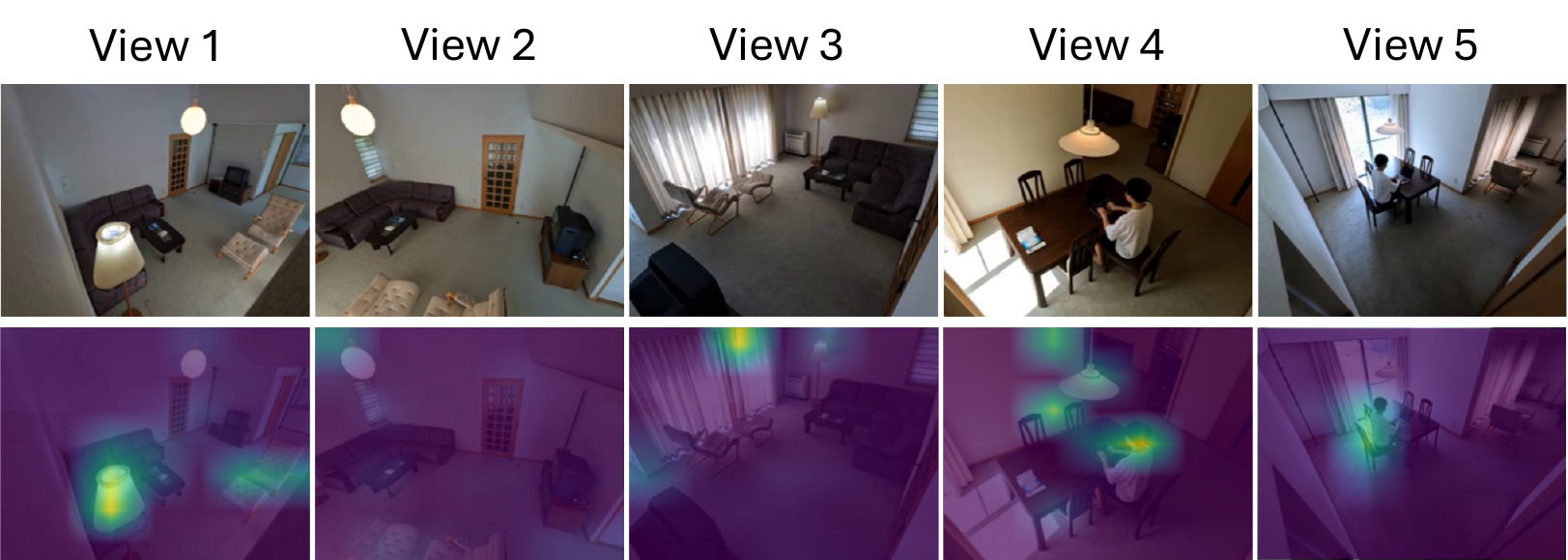}
    \label{fig:attention_map_1}
    }
    \vspace{-5pt}
    \hfill
    \subfigure[Temporal sequence of video frames (top row) and attention heatmaps (bottom row) highlighting action-relevant regions over time.]{
    \centering
    \includegraphics[width=0.47\textwidth]{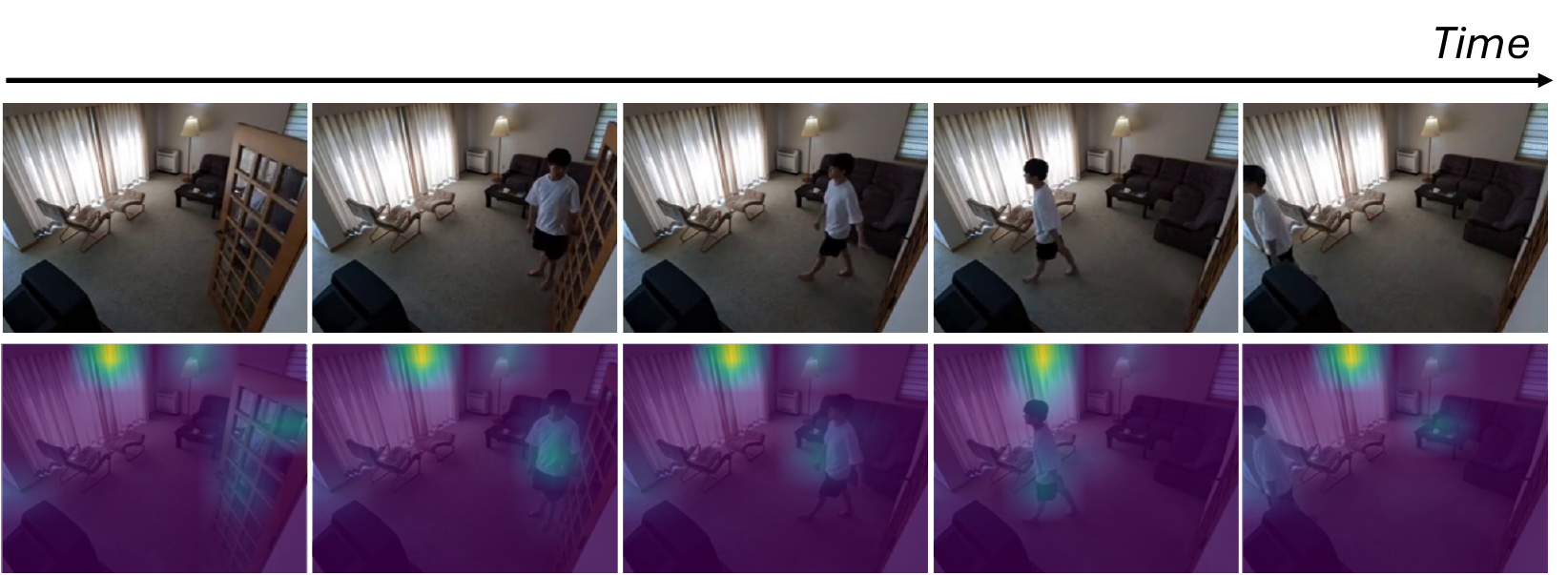}
    \label{fig:attention_map_2}}
    \caption{
    Visualization of multi-view and temporal attention heatmaps from the Shared Visual Encoder on the MultiSensor-Home dataset.}
    \label{fig:attention_map}
    \vspace{-5pt}
\end{figure}

Figure~\ref{fig:attention_map} illustrates the attention heatmaps from the Shared Visual Encoder on the MultiSensor-Home dataset, demonstrating its ability to identify action-relevant regions across both spatial and temporal dimensions.
Figure~\ref{fig:attention_map_1} shows multi-view inputs alongside the attention heatmaps, revealing that the model effectively focused on key objects such as lamps, curtains, desks, and human presence, which are critical for action recognition. 
Figure~\ref{fig:attention_map_2} visualizes the temporal attention maps, demonstrating how the model tracked significant changes in regions-of-interest over time, such as human movement and interaction with objects.

\begin{figure}[t]
    \centering
    \includegraphics[width=0.46\textwidth]{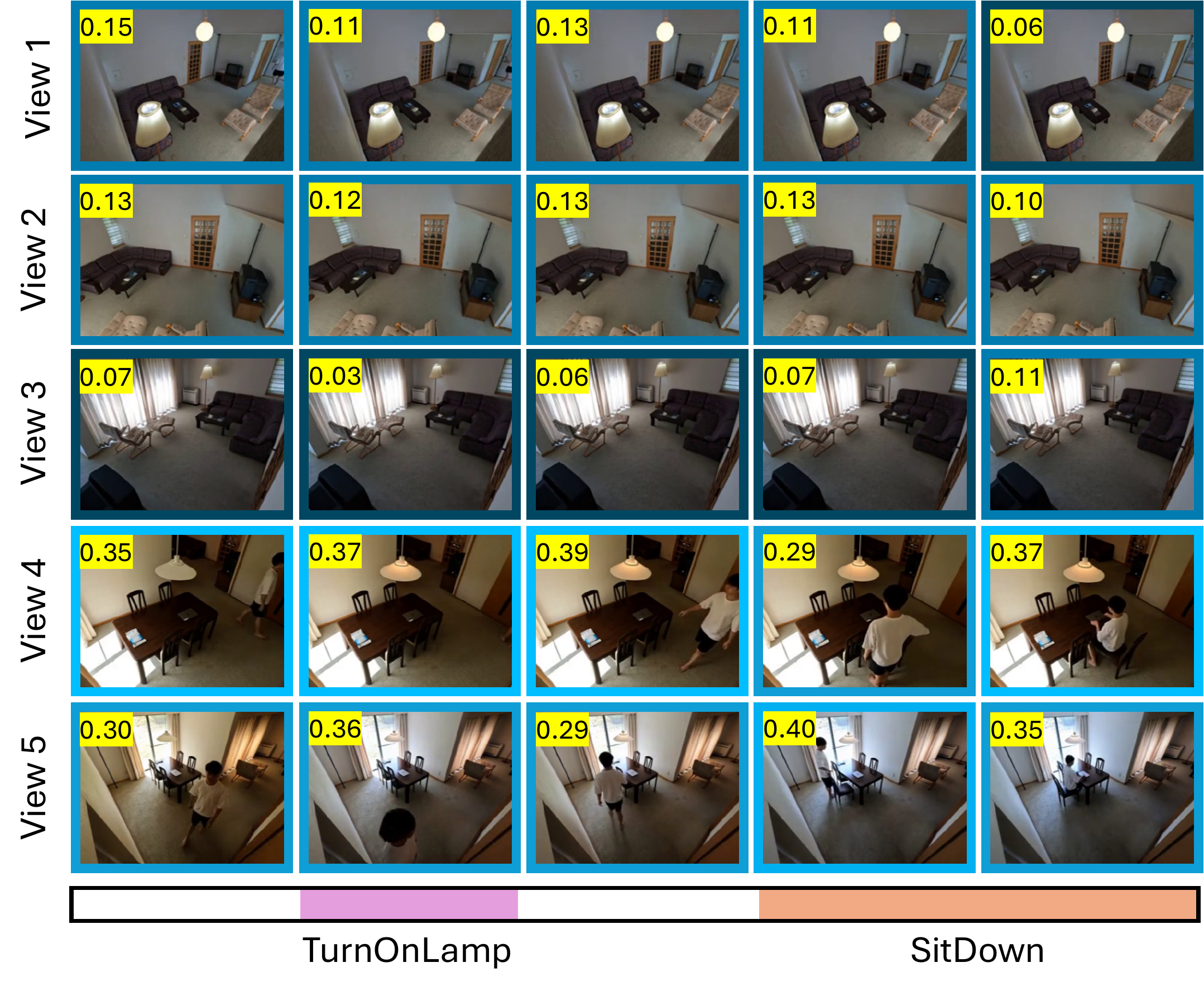}
    \caption{Attention scores from the Transformer-based Sensor Fusion across multiple views on the MultiSensor-Home dataset.}
    \label{fig:attention_scores_TSF}
    \vspace{-5pt}
\end{figure}

Additionally, Figure~\ref{fig:attention_scores_TSF} visualizes the attention scores from the Transformer-based Sensor Fusion across multiple views on the MultiSensor-Home dataset, highlighting the ability of the model to dynamically assign importance to different views based on their relevance to specific actions, such as ``TurnOnLamp'' and ``SitDown''.
The attention scores demonstrate the model's focus on the most informative perspectives (i.e., Views 4 and 5), further validating the effectiveness of the proposed fusion mechanism.

\subsection{Ablation Studies}
To evaluate the effectiveness of individual components in the proposed MultiTSF, we conduct a series of ablation studies on the MultiSensor-Home dataset.

\subsubsection{Effectiveness of Loss Components}
Table~\ref{table:loss_components} shows the contributions of each loss component. 
The results demonstrate that the full model (row 1) achieved the highest performance in both the uni-modal and multi-modal settings. 
Notably, excluding \(\mathcal{L_\text{F}}\) (row 2) or \(\mathcal{L_\text{S}}\) (row 3) lead to noticeable performance degradation, emphasizing the importance of frame-level and sequence-level supervision. 
The removal of \(\mathcal{L_\text{H}}\) (rows 4, 5, and 6) resulted in a significant performance drop, particularly in the multi-modal setting, underscoring its essential role in guiding the model to focus on human-centric regions for effective action recognition. 
Additionally, replacing \(\mathcal{L_\text{F}}\) and \(\mathcal{L_\text{S}}\) with Cross Entropy loss (row 7) further reduced accuracy, highlighting the superiority of Two-Way loss in handling class imbalance effectively.

\subsubsection{Sensor Fusion Strategies}
We compare the proposed Transformer-based Sensor Fusion with other widely used fusion strategies, including max pooling, mean pooling, and concatenation. 
Table~\ref{table:device_fusion} shows that the Transformer-based approach achieved the best results across both uni-modal and multi-modal settings. 
While mean pooling performed competitively in certain metrics, its inability to adaptively model inter-device relationships limited its overall effectiveness. 
Similarly, concatenation captured features from multiple devices but lacked the capability to prioritize relevant views dynamically. 
In contrast, the proposed Transformer-based fusion report better accuracy by effectively modeling inter-device dependencies and assigning importance to the most relevant views, resulting in superior performance across all metrics.

\begin{table}[t]
\centering
\caption{Effectiveness of loss components. The best and second-best results are highlighted in \textbf{bold} and \underline{underlined} text, respectively.} 
{
  \begin{tabular}{c|ccc|cc|cc}
    \toprule
    \multirowcell{2}[-2pt][l]{Row} & \multirowcell{2}[-2pt][l]{$\mathcal{L_\text{H}}$}  & \multirowcell{2}[-2pt][l]{$\mathcal{L_\text{S}}$} & \multirowcell{2}[-2pt][l]{$\mathcal{L_\text{F}}$} & \multicolumn{2}{c}{Uni-modal (Visual)} & \multicolumn{2}{c}{Multi-modal}  \\
    \cmidrule(lr){5-8}
     & & & & mAP${_C}$ & mAP${_S}$ & mAP${_C}$ & mAP${_S}$ \\
    \midrule
    1 & $\checkmark$ & $\checkmark$ & $\checkmark$ & \textbf{75.07} & \textbf{87.31} & \textbf{76.12} & \textbf{91.45} \\
    2 & $\checkmark$ & $\checkmark$ & & 61.17 & 84.22 & 64.48 & 87.91 \\
    3 & $\checkmark$ & & $\checkmark$ & 67.11 & 83.82 & 67.26 & 82.40 \\
    4 & & $\checkmark$ & $\checkmark$ & \underline{71.93} & \underline{86.95} & \underline{72.48} & \underline{90.80} \\
    5 & & $\checkmark$ & & 57.45 & 83.49 & 58.16 & 82.35 \\
    6 & & & $\checkmark$ & 66.43 & 81.65 & 67.18 & 83.84 \\
    \midrule
    
    \multicolumn{8}{c}{Change $\mathcal{L_\text{S}}$ and $\mathcal{L_\text{F}}$ to Cross Entropy loss} \\ \midrule
    7 & $\checkmark$ & $\checkmark$ & $\checkmark$ & 67.48 & 83.93 & 69.57 & 86.53 \\
    \bottomrule
  \end{tabular}
  \label{table:loss_components}
}
\end{table}

\begin{table}[t]
\centering
\caption{Effectiveness of sensor fusion strategies. 
The best and second-best results are highlighted in \textbf{bold} and \underline{underlined} text, respectively.} 
{
  \begin{tabular}{c|cc|cc}
    \toprule
     \multirowcell{2}[-2pt][l]{Sensor Fusion Strategies} & \multicolumn{2}{c}{Uni-modal (Visual)} & \multicolumn{2}{c}{Multi-modal}  \\
    \cmidrule(lr){2-5}
     & mAP${_C}$ & mAP${_S}$ & mAP${_C}$ & mAP${_S}$ \\
    \midrule
    Max Pooling & 69.13 & 87.42 & 72.24  & 88.83 \\
    Mean Pooling & \underline{71.86} & \textbf{91.20} & \underline{72.61} & 91.17 \\
    Concatenate & 71.69 & \underline{89.07} & 71.42 & \underline{91.18} \\
    Transformer (Proposed) & \textbf{75.07} & 87.31 & \textbf{76.12} & \textbf{91.45} \\
    \bottomrule
  \end{tabular}
  \label{table:device_fusion}
}
\vspace{-5pt}
\end{table}

\section{Conclusion}
\label{sec:conclusion}

In this study, we proposed the MultiSensor-Home dataset and the Multi-modal Multi-view Transformer-based Sensor Fusion (MultiTSF) method to address challenges in multi-modal multi-view action recognition. 
The MultiSensor-Home dataset provides a robust benchmark with untrimmed, multi-view sensors and detailed frame-level annotations.
The proposed MultiTSF method demonstrated its superiority over state-of-the-art methods on the proposed MultiSensor-Home and MM-Office~\cite{yasuda2022multi} datasets by effectively fusing audio and visual modalities through a Transformer-based mechanism and leveraging human detection for enhanced spatial learning. 
The results highlight significant improvements in mAP\(_C\) (macro-averaged metric) and mAP\(_S\) (micro-averaged metric), establishing MultiTSF as a competitive solution for real-world applications.
In future work, we aim to optimize models for deployment and expand the MultiSensor-Home dataset to include more home environments.


\section*{Acknowledgment}
This work was partly supported by JSPS KAKENHI JP21H03519 and JP24H00733. 
The computation was carried out using the General Projects on supercomputer ``Flow'' at Information Technology Center, Nagoya University.


\section*{Ethical Impact Statement}
In this paper, we introduced the MultiSensor-Home dataset and proposed the MultiTSF method for multi-modal multi-view action recognition. 
Below, we outline the ethical considerations and potential societal impacts of this study.

\subsection*{Data Collection and Privacy}
The MultiSensor-Home dataset comprises video and audio recordings captured in an indoor environment using multiple synchronized sensors. 
The dataset was collected with explicit consent from participants, ensuring compliance with ethical guidelines on data privacy and protection. 
No personally identifiable information is included in the dataset, and all participants were informed about the purpose of data collection, usage, and potential applications. 
However, anonymizing the videos (e.g., through blurring or masking) is not feasible, as it would compromise the dataset's utility and the validity of the research findings.

\subsection*{Bias and Fairness}
The MultiSensor-Home dataset was designed to include diverse scenarios, such as variations in lighting conditions, clothing, and environmental settings, to ensure robustness across different contexts. 
However, potential biases may persist due to the data being collected from a single home environment. 
Future work should aim to expand the dataset to include a wider range of environmental conditions, improving fairness and generalizability.

\subsection*{Potential Misuse and Societal Impact}
The proposed MultiTSF method aims to improve action recognition performance in various applications, including smart homes, assistive technologies, and human-computer interaction. 
However, like any computer vision technology, there is a risk of misuse, such as unauthorized surveillance or privacy-intrusive applications. 
To mitigate these risks, we encourage the responsible and ethical use of the proposed dataset and method in accordance with legal and ethical frameworks. 
The proposed dataset and method are intended solely for research purposes, and commercial usage should adhere to ethical AI principles.

\subsection*{Transparency and Reproducibility}
To promote transparency and reproducibility, we provide detailed descriptions of the proposed MultiSensor-Home dataset and the MultiTSF method. 
The dataset and method implementation will be made publicly available under appropriate licensing to ensure responsible research practices and facilitate further advancements in the field.

\subsection*{Conclusion}
In this paper, we contribute to the advancement of multi-modal multi-view action recognition while addressing ethical concerns related to privacy, bias, and potential misuse. 
We emphasize the importance of ethical AI practices and encourage future research to continue improving fairness, privacy safeguards, and responsible deployment of AI-driven action recognition technologies.


{\small
\bibliographystyle{ieee}
\bibliography{egbib}
}

\end{document}